\documentclass[conference,letterpaper,10pt]{IEEEtran}
\IEEEoverridecommandlockouts

\usepackage[T1]{fontenc}
\usepackage[utf8]{inputenc}
\usepackage{lmodern}
\usepackage{amsmath,amssymb,amsfonts,bm,mathtools}
\usepackage{booktabs,tabularx,array,multirow}
\usepackage{graphicx}
\usepackage[dvipsnames]{xcolor}
\usepackage{tikz}
\usetikzlibrary{arrows.meta,positioning,calc,fit,backgrounds}
\usepackage{algorithm}
\usepackage{algpseudocode}
\usepackage{cite}
\usepackage{microtype}
\usepackage[hidelinks]{hyperref}
\usepackage[capitalise,noabbrev]{cleveref}
\usepackage{balance}
\usepackage{cuted}
\usepackage{caption}

\definecolor{DeepInk}{HTML}{213F52}
\definecolor{SoftTeal}{HTML}{DDEEEB}
\definecolor{SoftOrange}{HTML}{FAEBDD}
\definecolor{AccentTeal}{HTML}{217668}
\definecolor{AccentOrange}{HTML}{AF6538}
\definecolor{GreyLine}{HTML}{A8B1B6}

\IfFileExists{computed_results.tex}{\input{computed_results.tex}}{}
\IfFileExists{computed_transfer.tex}{\input{computed_transfer.tex}}{}

\title{\LARGE\bfseries
DiffGR00T: Execution-Aware Differentiable Training for Humanoid Whole-Body Vision-Language-Action Models}

\author{\IEEEauthorblockN{Sichen Wu, Qian Sun, Mengfan Liu,
Weikai Qin, Bike Zhu, and Hesheng Wang\(^{\dagger}\)}
\thanks{Sichen Wu, Qian Sun, Mengfan Liu, Weikai Qin, and Bike Zhu are with
the Department of Automation, Shanghai Jiao Tong University, Shanghai 200240,
China.\par
\(^{\dagger}\)Hesheng Wang is with the Department of Automation, Shanghai Jiao
Tong University, Shanghai 200240, China.}}

\begin{document}
\maketitle

\begin{strip}
\vspace*{-80pt}
\centering
\includegraphics[width=\textwidth]{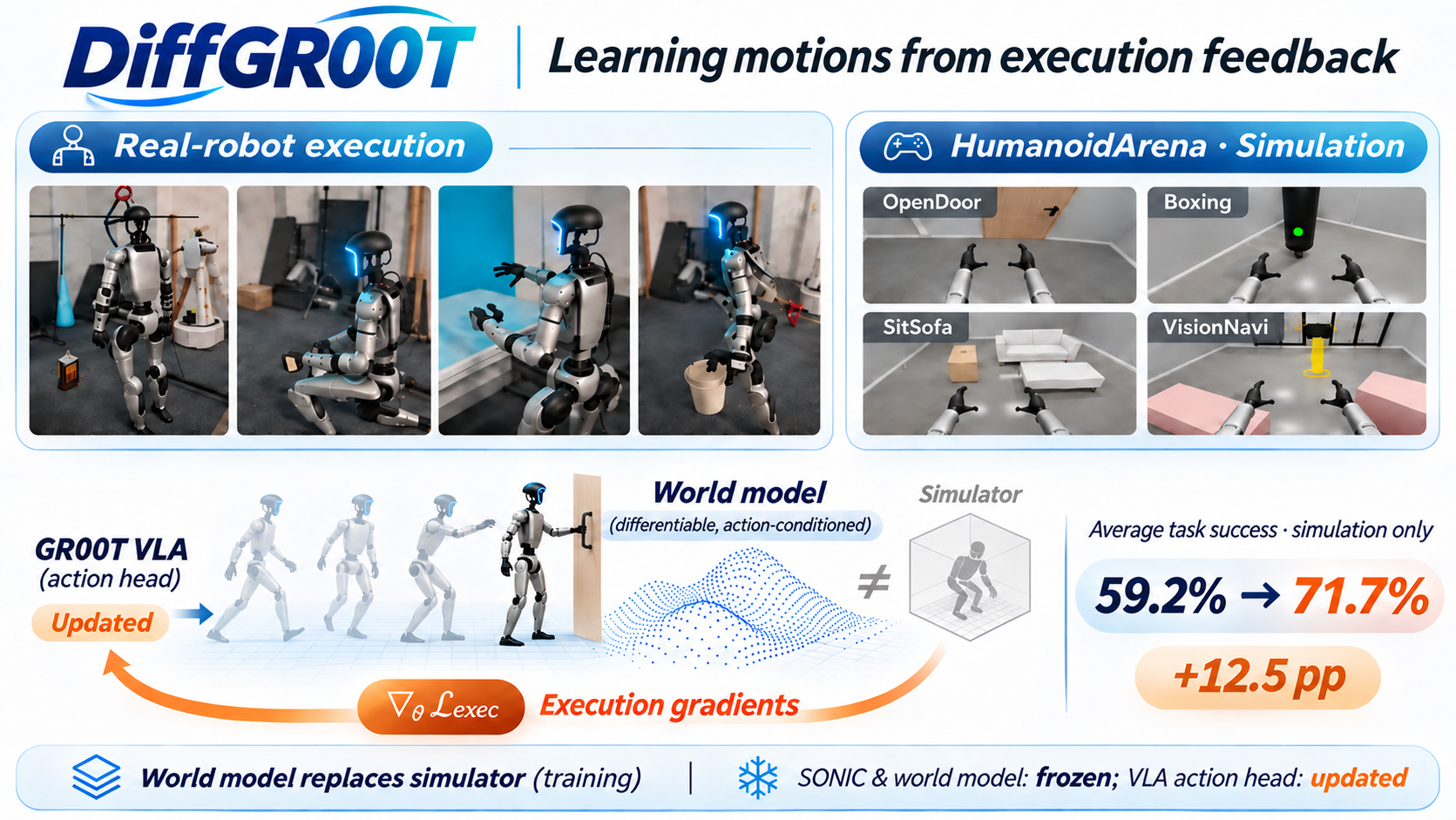}
\captionof{figure}{\textbf{DiffGR00T teaser.} The VLA action head is updated
while SONIC and the world model are frozen; the world model replaces the
simulator during training. Simulation-only average task success rises from 59.2\% to
71.7\%.}
\label{fig:teaser}
\end{strip}

\begin{abstract}
Vision-language-action (VLA) models have achieved strong results in arm-centric manipulation, yet humanoid whole-body control (WBC) requires dynamic balance and tightly coupled full-body coordination. Existing hierarchical humanoid VLAs typically decouple motion generation from physical execution, leaving tracking and balance errors unable to directly optimize the VLA generator. We propose \textbf{DiffGR00T}, an end-to-end differentiable training framework that couples GR00T N1.7 with SONIC, a pretrained whole-body behavior foundation model. To enable execution-aware differentiable training, we replace the non-differentiable simulator with a learned, action-conditioned world model, allowing multi-step body-execution losses to backpropagate through the frozen whole-body controller and the flow-matching sampler directly into the VLA action head. On four HumanoidArena simulation tasks, DiffGR00T improves average task success by 12.5 percentage points over the supervised base VLA (59.2\% $\rightarrow$ 71.7\%), demonstrating that execution-aware gradients effectively align high-level motion generation with closed-loop whole-body dynamics.
\end{abstract}
% \begin{abstract}
% Vision-language-action (VLA) models have advanced arm-centric manipulation, yet humanoid whole-body control (WBC) tasks demand coordinated control across the entire body and dynamic stability. Existing humanoid VLAs typically optimize motion generation separately from physical execution, leaving execution errors disconnected from VLA learning. We propose DiffGR00T, an end-to-end co-training framework that couples GR00T N1.7 with SONIC, a released whole-body behavior foundation model. During training, gradients are routed through a differentiable, action-conditioned world model
% instead of through the simulator, so that the execution error of each short rollout can be
% backpropagated through the whole-body controller and the flow-matching sampler into the action head. On HumanoidArena, \Method{}
% improves average task success by 12.5 points over the base VLA trained without co-training
% (59.2\% $\rightarrow$ 71.7\%), showing that execution-aware gradients can improve how a humanoid VLA acts on the body it controls.
% \end{abstract}

\section{Introduction}
\label{sec:intro}

Vision-language-action (VLA) models have made remarkable progress in
arm-centric manipulation, where predicted end-effector poses or
joint-space targets can be executed through low-level servo controllers
\cite{black2024pi0,chen2026dial}. Extending these capabilities to
humanoid robots, however, introduces fundamentally different control
challenges. Unlike fixed-base manipulators, humanoids exhibit
underactuated floating-base dynamics in which locomotion, manipulation,
and posture regulation must be coordinated under changing contact and
balance constraints. Directly predicting joint targets does not
guarantee stable whole-body execution, while learning high-frequency
stabilization directly inside a large VLA remains difficult.
Consequently, many humanoid systems adopt a hierarchical architecture in
which a VLA generates high-level motion intentions and a pretrained
behavior foundation model (BFM) or whole-body controller realizes them
through proprioceptive feedback.

This hierarchical paradigm has driven substantial progress in humanoid
loco-manipulation. SONIC establishes a generalizable whole-body motion
foundation through a shared latent interface, while systems such as
WholeBodyVLA and OpenHLM pair task-conditioned action generation with
pretrained motion controllers
\cite{luo2025sonic,jiang2025wholebodyvla,hu2026openhlm}. Because these
architectures train the VLA to emit motion references and the controller
to track them independently, the controller provides reactive
whole-body stabilization at deployment, yet an \emph{action--execution
training gap} remains: tracking and balance errors have no direct
gradient path back into the VLA's motion-generation objective. Closing
this gap via end-to-end differentiable training requires establishing a
differentiable gradient pathway across the policy--controller interface.

Bridging this gap is non-trivial because robust whole-body control and
differentiable optimization have traditionally followed separate
training paradigms. Humanoid motion trackers are commonly trained with
model-free reinforcement learning, where physical execution feedback
arrives as scalar rewards rather than differentiable state transitions.
Moreover, contact-rich humanoid dynamics involve non-smooth support
switches, making direct differentiation through conventional rigid-body
simulators numerically brittle and computationally expensive. Although
differentiable physics methods such as DiffMimic have explored
analytical motion-imitation gradients \cite{ren2023diffmimic}, scaling
such optimization to high-dimensional whole-body control and large VLA
generators remains computationally prohibitive.

An alternative to differentiating through the simulator is to
differentiate through a learned world model. Fitted to state transitions
collected from the simulator or the robot, an action-conditioned neural
world model is differentiable everywhere---including across contact
transitions---allowing execution objectives to be backpropagated through
predicted states without invoking a contact solver, while short rollout
windows bound compounding prediction error.
SuperMimic~\cite{liao2026supermimic} demonstrates this paradigm for
humanoid tracking by training a tracking policy with short-horizon
backpropagation through time (BPTT) on a learned world model, and in a
velocity-tracking experiment further routes gradients through a shared
decoder into a lightweight goal encoder. Extending this
execution-gradient pathway across a frozen token-conditioned controller
and a multi-step flow-matching sampler into a humanoid VLA, however,
remains largely unexplored.

We present \textbf{DiffGR00T}, an \emph{end-to-end differentiable
training} framework that couples GR00T~N1.7 with the pretrained SONIC
behavior foundation model. We first fine-tune the GR00T~N1.7 action head
on HumanoidArena demonstrations carrying recorded SONIC motion
tokens~\cite{nvidia2026groot,wang2026humanoidarena} to align the VLA
with SONIC's latent motion interface. During \emph{execution-aware
differentiable training}, a learned action-conditioned world model
supplies multi-step state transitions to evaluate a body-execution loss,
allowing execution gradients to propagate through the world model, the
frozen SONIC decoder, and the flow-matching sampler directly into the
VLA action head.

Our mechanism exploits the temporal structure of the VLA's action
chunks. The flow-matching VLA generates a 40-step motion-token sequence,
of which the first 20 tokens are executed in closed loop. At each
50\,Hz control step, the frozen SONIC decoder maps the current token and
proprioceptive history to 29 joint targets, and the world model predicts
the resulting robot state, which is fed back as proprioception for the
next step. Multi-step body-execution losses are then backpropagated
through a ten-step rollout window, the frozen decoder, and the four-step
flow-matching sampler to update only the VLA action head (via LoRA adapters and
the output projection), while the world model and the training-only critic
remain fixed throughout. On four HumanoidArena simulation tasks, DiffGR00T raises
average task success by 12.5 percentage points over the supervised base
VLA (59.2\% $\rightarrow$ 71.7\%).

Our main contributions are as follows:

\begin{enumerate}

\item \textbf{A differentiable world model for whole-body execution.}
We learn an action-conditioned dynamics model for a 29-DoF humanoid from
controller-excitation replay and a small amount of closed-loop policy
data, holding its parameters fixed during execution-aware differentiable
training to serve as a differentiable surrogate for the simulator over
short rollouts.

\item \textbf{Execution gradients into a flow-matching VLA.}
We backpropagate multi-step body-execution losses through the learned
world model, the frozen SONIC decoder, and the four-step flow-matching
sampler to update the VLA action head, while keeping the vision-language
backbone, controller, world model, and training-only critic fixed.

\item \textbf{Execution-aligned supervision.}
We identify the key supervision terms required for effective humanoid
execution feedback---root position and drift to curb per-window
drift, transition-frame weighting for task-phase changes, and
speed-coverage window sampling to retain locomotion in the gradient
budget---calibrated to maintain scale consistency across objectives.

\end{enumerate}
\section{Related Work}
\label{sec:related}

\subsection{Flow-Matching Policies and Policy Improvement}
\label{sec:related_fm}

$\pi_0$ combines a pretrained vision-language backbone with a flow-matching action expert to generate continuous action chunks \cite{black2024pi0}. Building on this paradigm, recent work explores alternative gradient pathways for supervised policy learning: DIAL links the VLM and a reactive action expert through predicted visual intent, backpropagating the supervised flow-matching loss through this latent intent into trainable VLM components during joint fine-tuning \cite{chen2026dial}; VITA, a visuomotor policy mapping image encodings to latent actions, differentiates a supervised action-reconstruction loss directly through the Euler steps of the flow-matching ODE solver \cite{gao2025vita}.

A complementary line of research incorporates robot execution experience into flow-matching policy updates. RECAP fits a value function to execution rollouts and conditions supervised VLA updates on value-derived advantage indicators \cite{pi2025recap}, whereas Q-VGM learns an action-chunk critic from rollout rewards and converts critic gradients with respect to denoised action chunks into local velocity targets, avoiding backpropagation through the full denoising trajectory \cite{wang2026qvgm}. While these methods leverage ODE differentiation, advantage conditioning, or value guidance, none propagates a multi-step body-execution objective through a whole-body controller and learned action-conditioned dynamics into the flow-matching generator.

\subsection{Humanoid Whole-Body Action Generation and Control}
\label{sec:related_humanoid}

Humanoid loco-manipulation systems span diverse interfaces between high-level action generation and low-level control. SONIC learns a generalizable whole-body motion tracker from motion-capture data and demonstrates VLA-driven loco-manipulation with GR00T N1.5 through a shared motion-token interface \cite{luo2025sonic}. HumanoidArena benchmarks hierarchical policy--tracker pairings and documents a direct prediction mode in which a VLA outputs 64-dimensional SONIC body latents \cite{wang2026humanoidarena}, providing the interface and benchmark setting adopted in this work. Other hierarchical frameworks rely on different intermediate representations: WholeBodyVLA learns locomotion and manipulation latents from action-free egocentric video and decodes them into upper-body joint targets alongside locomotion commands executed by a reinforcement-learning controller \cite{jiang2025wholebodyvla}, while OpenHLM combines joint-space teleoperation with heterogeneous data co-training to generate whole-body joint references tracked by an independent low-level policy \cite{hu2026openhlm}.

Beyond single-stage reference generation, HAF-VLA produces cumulative whole-body commands via stage-wise flow matching and cross-stage conditioning, with existing reinforcement-learning controllers executing locomotion and waist commands \cite{gu2026hafvla}. Departing from explicit reference tracking, SENTINEL clones a simulator-trained whole-body controller into a language-and-proprioception policy that predicts low-level action targets directly, then freezes that policy and trains a residual action head with PPO \cite{wang2025sentinel}. Across these interfaces---shared motion tokens, joint-space references, staged commands, and direct low-level targets---execution discrepancies are absorbed by downstream trackers or a separate residual head rather than backpropagated through the controller to update the high-level VLA generator.

\subsection{Differentiable Whole-Body Control and Execution-Aware Learning}
\label{sec:related_diff}

Propagating physical execution gradients directly into policy learning has progressed from differentiable physics to learned dynamics surrogates. DiffMimic trains a humanoid motion-tracking policy by differentiating imitation losses through a differentiable rigid-body simulator \cite{ren2023diffmimic}. To avoid differentiating through contact solvers, ControlVAE optimizes task policies for simulated characters by backpropagating task losses through a frozen latent skill controller and a learned world model \cite{yao2022controlvae}. MoConVQ similarly uses a learned world model to train discrete motion representations and their decoder, while training its text-conditioned generator separately via token prediction \cite{yao2024moconvq}.

Closest to our humanoid setting are SuperMimic and HWAM. SuperMimic fits a differentiable world model from simulator transitions and trains a humanoid tracking policy via short-horizon imagined rollouts; in a downstream velocity-tracking experiment, execution gradients also pass through the world model and the shared decoder (with decoder weights frozen on that branch) into a velocity goal encoder \cite{liao2026supermimic}. HWAM incorporates physical execution by jointly modeling kinematic references and aligned measured post-execution states, coupled with auxiliary objectives that predict future video from joint trajectories and reconstruct joint trajectories from visual transitions \cite{yang2026hwam}.

In contrast, DiffGR00T performs execution-aware differentiable training on a flow-matching humanoid VLA: multi-step body-execution losses are backpropagated through a short rollout of learned action-conditioned dynamics, the frozen SONIC decoder, and the four-step flow-matching sampler to update only the VLA action head (via LoRA adapters and the output projection), leaving both the VLM backbone and the controller fixed. We evaluate this pathway along three axes: gradient connectivity into the flow head, directional agreement with simulator returns, and closed-loop task performance in simulation.

\section{Method}
\label{sec:method}
\begin{figure*}[!t]
\centering
\includegraphics[width=\textwidth]{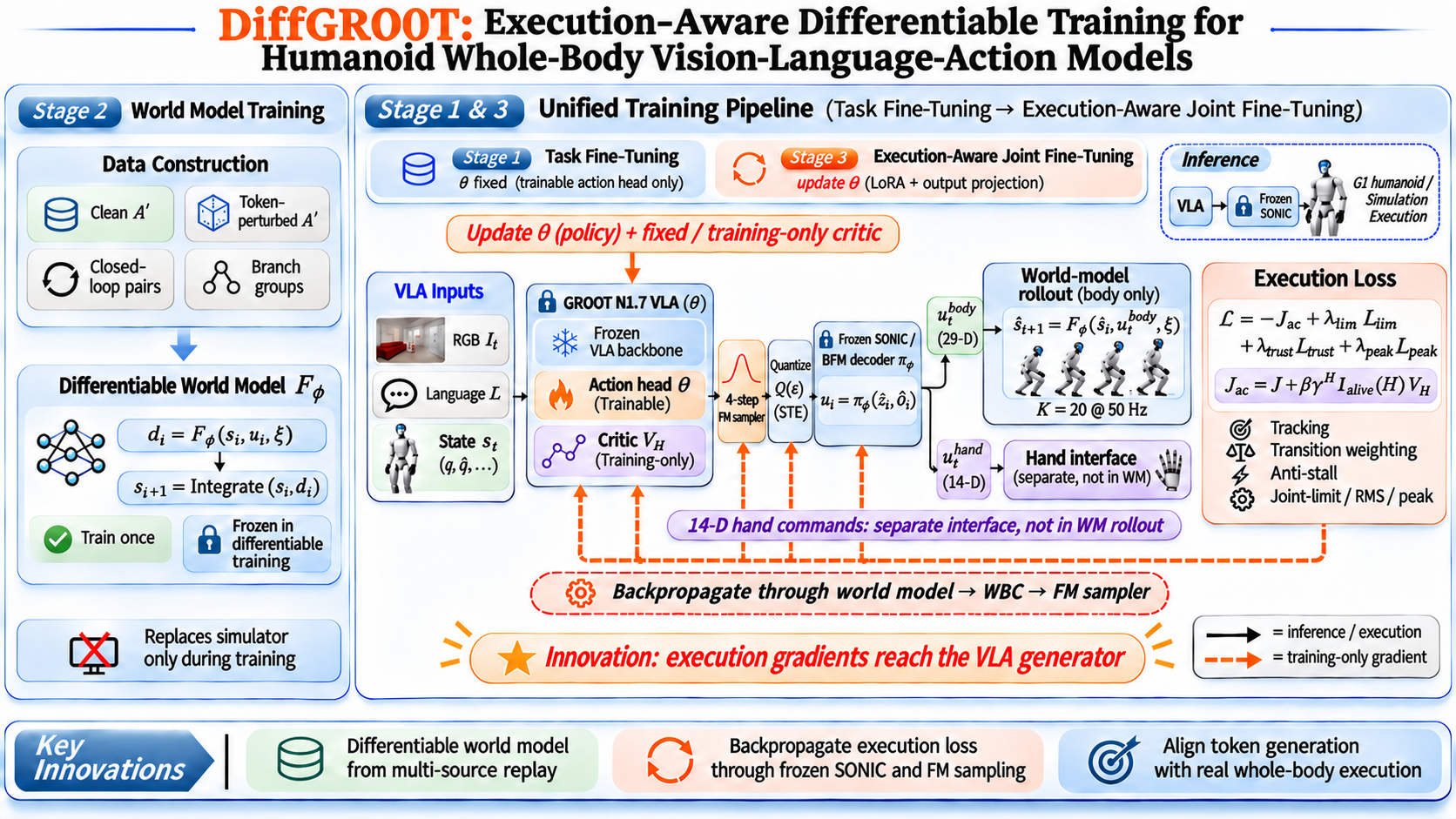}
\caption{\textbf{DiffGR00T architecture.} The GR00T N1.7 VLA action head is
trained; the VLA backbone, the SONIC decoder, the action-conditioned world model,
and the training-only critic are frozen. The execution loss is backpropagated
through the world model, the SONIC decoder, and the four-step flow-matching
sampler. The critic supplies a fixed tail value and is not part of the deployed
controller. The world model is trained once from multi-source replay and replaces
the simulator only during training; it is not used at inference.}
\label{fig:architecture}
\end{figure*}

\subsection{Overall Architecture and Control Interface}
\label{sec:overview}

As illustrated in \Cref{fig:architecture}, the framework consists of three coupled
components: a flow-matching VLA, a frozen whole-body foundation model SONIC, and
an action-conditioned differentiable world model. The VLA maps egocentric
observations, a language instruction, and robot state to a chunk of motion tokens;
the BFM decodes each token, together with a proprioceptive history, into
joint-level commands; and the world model predicts how the robot state evolves
under those commands.

The architecture is shared by all tasks and has two distinct execution modes. At
deployment, only the VLA, the frozen SONIC decoder, and the hand interface run:
the VLA produces an action chunk, SONIC decodes its body tokens and
proprioceptive history into 29-dimensional joint commands, and the separate hand
interface executes the 14 finger commands. The world model and the training-only
critic are absent, so the learned surrogate adds no inference cost.

During training, the simulator is replaced by the action-conditioned world model.
For the body branch, the same chunk is passed through the frozen SONIC decoder and
the world model, and the resulting execution cost is backpropagated through the
world model, the frozen decoder, and the four-step flow-matching sampler to the
VLA action head. Because the frozen modules remain in the backward graph, their
parameters do not change but their Jacobians do. This frozen-but-differentiable
path is the central architectural difference from standard VLA fine-tuning. The
world model is trained once from multi-source replay and is never deployed; the
critic is training-only and summarizes the tail of the rollout.

The VLA generates a 40-step chunk through a four-step flow-matching sampler. At
time \(t\), the chunk is
\[
A_t=[Z_t,H_t]\in\mathbb{R}^{40\times78},
\]
where the body part \(Z_t\in\mathbb{R}^{40\times64}\) collects
SONIC-compatible motion tokens \(z_{t+i|t}\in\mathbb{R}^{64}\) for
\(i=0,\ldots,39\), and the hand part \(H_t\in\mathbb{R}^{40\times14}\) collects
seven finger commands per hand for each step. Each step is therefore a
78-dimensional vector: 64 body-token dimensions plus 14 finger dimensions.

For each body step \(j=t+i\), the frozen body foundation model
\(\pi_\psi\) takes the token \(z_j\) and the ten-frame proprioceptive history
\(O_j\in\mathbb{R}^{10\times93}\), and outputs a 29-dimensional normalized body
action
\begin{equation}
u_j=\pi_\psi(z_j,O_j).
\label{eq:bfm}
\end{equation}
The 29 entries of \(u_j\) correspond one-to-one to the 29 body joints; they are
normalized action coordinates, not joint angles yet. Each history frame in
\(O_j\) stacks 29 joint-position offsets from the default configuration,
29 joint velocities, 3 body angular-velocity components, 3 projected-gravity
components, and the 29-dimensional previous action
(\(29+29+3+3+29=93\)).

The normalized action is then converted into joint-position targets,
\begin{equation}
q_j^{\mathrm{cmd}}=q^{\mathrm{default}}+\boldsymbol{\alpha}\odot u_j,
\label{eq:joint_target}
\end{equation}
where \(q^{\mathrm{default}}\in\mathbb{R}^{29}\) is the nominal joint
configuration, \(\boldsymbol{\alpha}\in\mathbb{R}^{29}\) is the per-joint action
scale, and \(\odot\) is elementwise multiplication; low-level PD controllers
track \(q_j^{\mathrm{cmd}}\). The 14 finger commands are collapsed to
HumanoidArena's binary open--close convention and executed through a separate
hand interface, so they do not enter the 29-dimensional body action.

The VLA is initialized from GR00T~N1.7 and fine-tuned on task demonstrations that
carry recorded motion tokens, with the vision-language backbone and the BFM held
fixed, so that only the action head is trained. This stage yields task-conditioned
token chunks, but it never observes the physical consequence of executing them;
Sec.~\ref{sec:joint_training} turns that consequence into a gradient.

\subsection{Differentiable World Model}
\label{sec:world_model}

In training the simulator is replaced by an action-conditioned world model
$F_\phi$ that predicts the transition caused by a body action. The model predicts
a one-step state increment rather than an absolute next state: an absolute target
is dominated by the robot's global pose, while the transition over a 20\,ms step
is small and approximately stationary. The network therefore outputs $d_t$, and
the next state is recovered by integration. Let \(\mathcal{I}\) denote the
integration operator that maps a state and a predicted increment to the next
state. Then
\begin{equation}
\begin{aligned}
d_t&=F_\phi(s_t,a_t,\xi)\in\mathbb{R}^{70},\\
\hat{s}_{t+1}&=\mathcal{I}(s_t,d_t),
\end{aligned}
\label{eq:world_model}
\end{equation}
where $s_t\in\mathbb{R}^{71}$ denotes the root pose (position and a unit
quaternion), the 29 joint positions, the root linear and angular velocities, and
the 29 joint velocities. The action $a_t\in\mathbb{R}^{29}$ is the normalized body
action produced by the BFM, and $\xi\in\mathbb{R}^{33}$ is a simulation-context
vector (friction, encoder bias, and center-of-mass offset) held at nominal values
at evaluation. The state is 71-dimensional and the increment is 70-dimensional
because orientation is stored as a quaternion in $s_t$ but predicted as a 3-D
rotation vector in $d_t$.

Integration proceeds in two stages. First, the velocity increments produce the
next root linear velocity, root angular velocity, and joint velocities. Second,
these updated velocities advance the root position, root orientation, and joint
angles over the 20\,ms step. The predicted residual increments
$\Delta p,\Delta\theta,\Delta q$ are added as local corrections on top of this
velocity-based update. For orientation, the 3-D rotation vector is converted into
a rotation through an exponential map and composed with the current orientation.
Root-related increments are expressed in the current heading frame, so the target
does not depend on the robot's global position or yaw. The increment is therefore
small and frame-local, while the full next state is recovered by integration.

The action enters both through the MLP input and through a residual action head,
\begin{equation}
\begin{aligned}
F_\phi(s_t,a_t,\xi)={}&\mathrm{MLP}_\phi(s_t,a_t,\xi)\\
&+G_\phi(s_t)\,(a_t-a_{\rm ref}),
\end{aligned}
\label{eq:bilinear}
\end{equation}
where $a_{\rm ref}$ is a reference action. The residual term gives the action a
direct, state-dependent path that can be supervised on branch groups sharing a
state but differing in action. The MLP retains nonlinear action dependence, while
$G_\phi$ captures a local linear correction around $a_{\rm ref}$. $G_\phi$ is
zero-initialized, so training starts from the MLP and learns the residual
correction. Held-out checks confirm that the learned action dependence exceeds
the noise floor.

The model is trained on four transition sources (Table~\ref{tab:wmdata}). We
write A$'$ for the controller-excitation replay corpus collected from the frozen
BFM; clean A$'$ and token-perturbed A$'$ are its two subsets. The first is clean
controller-excitation replay: the frozen BFM is driven with
reference tokens in simulation, and the resulting state--action--next-state
triples are recorded without the VLA in the loop. It supplies the nominal
tracking anchor and a broad transition pool. The second repeats the same
collection with temporally correlated Gaussian token noise of 0.02, 0.05, and
0.10, held fixed for one 20-step chunk; because the BFM executes the perturbed
token, the stored action is still the action that produced the transition. This
broadens coverage to off-trajectory and recovery states. Periodic resets every
28--37 steps, with up to 0.3\,m displacement, add further recovery states. We
perturb tokens rather than actions so the stored action still matches the
executed transition.

The third source is closed-loop pairs collected with the VLA in the loop. These
cover the deployment distribution, including contact and drift. The fourth is
branch groups: one state with four different token sequences, which provide
same-state action contrast for the residual action head and the gain term. The
reference motion library is not a dynamics source; it is used only to define
reference frames.

\begin{table}[t]
\centering
\footnotesize
\setlength{\tabcolsep}{4pt}
\renewcommand{\arraystretch}{1.05}
\caption{\textbf{World-model data mix.} Stage lengths 880 and 1650; the second
stage uses $K=20$ autoregressive windows. Shares are per batch.}
\label{tab:wmdata}
\begin{tabular}{@{}lccc@{}}
\toprule
Source & Stage 1 & Stage 2 & Role \\
\midrule
Clean A$'$ & 0.15 & 0.20 & nominal dynamics \\
Token-perturbed A$'$ & 0.28 & 0.45 & off-trajectory states \\
Closed-loop pairs & 0.27 & 0.35 & deployment distribution \\
Branch groups & 0.30 & 0.00 & action contrast \\
\bottomrule
\end{tabular}
\end{table}

Training proceeds in two stages. In stage 1 we use single-step transitions
($K=1$); in stage 2 we use $K=20$ autoregressive windows. For a rollout
$\hat{s}_0,\ldots,\hat{s}_K$ started from $s_0$, let \(R(q)\) and \(H(q)\) denote
the rotation matrix and heading frame of the root quaternion. Let \(SO(3)\)
denote the rotation group and \(\operatorname{Log}:SO(3)\to\mathbb{R}^3\) the
matrix logarithm, which maps a rotation matrix to its rotation vector. Then
\(\|\operatorname{Log}(R(q)^\top R(\hat q))\|_2\) is the geodesic angle between
the predicted and recorded orientations. Define the block-wise state error
\begin{equation}
\mathrm{Err}(\hat s,s)=
\begin{bmatrix}
H(q)^\top(\hat p-p)/\sigma_p\\
\operatorname{Log}\!\left(R(q)^\top R(\hat q)\right)/\sigma_R\\
(\hat q_{\rm joint}-q_{\rm joint})/\sigma_q\\
R(q)^\top(\hat v-v)/\sigma_v\\
R(q)^\top(\hat \omega-\omega)/\sigma_\omega\\
(\hat{\dot q}-\dot q)/\sigma_{\dot q}
\end{bmatrix},
\label{eq:state_err}
\end{equation}
where \(\hat p,\hat q,\hat v,\hat \omega,\hat q_{\rm joint},\hat{\dot q}\) are the
predicted position, quaternion, root linear velocity, root angular velocity,
joint positions, and joint velocities. The state loss is
\begin{equation}
\mathcal{L}_{\rm state}
=
\frac{1}{K}\sum_{j=1}^{K}
\left\|
\mathrm{Err}(\hat s_j,s_j)
\right\|_2^2,
\label{eq:wm_state_loss}
\end{equation}
where $\sigma_p,\sigma_R,\sigma_q,\sigma_v,\sigma_\omega,\sigma_{\dot q}$ are
the per-block normalizations. The orientation block is a geodesic quaternion
error, and positions and velocities are expressed in the frame-local convention.
This avoids mixing metres, radians, and quaternion components in a single raw
\(L_2\) norm.

Branch groups add a gain loss. For a group with $K_b=4$ branches including the
reference branch, let $\hat{s}^{(k)}$ and $s^{(k)}$ be the predicted and recorded
states after executing branch $k$. The gain loss matches the predicted and true
state differences relative to the first branch:
\begin{equation}
\mathcal{L}_{\rm gain}
=
\mathbb{E}_{\mathcal{B}}
\left[
\frac{1}{K_b}
\sum_{k=1}^{K_b-1}
\left\|
\frac{(\hat{s}^{(k)}-\hat{s}^{(0)})-(s^{(k)}-s^{(0)})}
{\sigma_\Delta}
\right\|_2^2
\right],
\label{eq:wm_gain_loss}
\end{equation}
where $\mathcal{B}$ denotes branch groups and $\sigma_\Delta$ is a per-block
scale for positions, orientation, joints, and velocities. The gain loss is
computed on the raw state increment in the recorded state coordinates. In the
collected branch groups, quaternion signs are consistent with the reference
branch, so no sign canonicalization is required in this setting. The total
objective is
\begin{equation}
\mathcal{L}_{\rm WM}
=
\mathcal{L}_{\rm state}
+
\lambda_g\,\mathcal{L}_{\rm gain},
\label{eq:wm_total_loss}
\end{equation}
with $\lambda_g=2.0$ and an EMA-adaptive loss mix over the four sources. The
branch share is used only in stage 1 (Table~\ref{tab:wmdata}). The model is
trained once, accepted only if multi-step error does not regress on held-out
windows, and kept frozen while the policy is updated. We do not treat it as a
precise simulator; it only needs to provide approximate execution gradients for
the policy update.

The world model is not updated on-policy. A refresh is not a single extra collection:
it must be repeated as the policy moves, and each round requires closed-loop VLA
inference across multiple tasks and many episodes, followed by world-model
retraining and a new gate. A 300-update run would therefore need multiple such
rounds, making the total cost far larger than the one-hour fine-tune. Repeated
refresh also makes the objective non-stationary and can degrade the action-effect
calibration; an online-update variant did so in earlier tests. We therefore train
the world model once and keep it fixed. Closed-loop pairs are a fixed calibration
source, not a real-time update stream, and the fixed world model makes the closed-loop
comparison attributable to the policy update.

\begin{table}[t]
\centering
\footnotesize
\setlength{\tabcolsep}{4pt}
\renewcommand{\arraystretch}{1.05}
\caption{\textbf{Cost of on-policy world-model refresh.} A refresh is a repeated
round, not a one-off: each round requires closed-loop VLA data across multiple
tasks, WM retraining, and a new gate. The fixed world model is used for all
reported numbers.}
\label{tab:refresh}
\begin{tabular}{@{}lcc@{}}
\toprule
 & Fixed & On-policy \\
\midrule
Closed-loop VLA data & -- & multi-task, many episodes \\
WM retrain & -- & per round \\
WM gate & -- & per round \\
Policy updates & 59--66 min & per round \\
Refresh rounds & 0 & multiple \\
Total for one run & $\approx$1 h & $\gg$1 h \\
\bottomrule
\end{tabular}
\end{table}

\subsection{Execution-Aware Differentiable Training}
\label{sec:joint_training}

Differentiable training preserves the computation graph from the chunk through the
BFM, the world model, and the sampler, so that an objective defined on imagined
execution reaches the generator. Everything except the VLA's action head is held
fixed during the update: the BFM, the SONIC encoder, the world model, and the
training-only critic keep their parameters. The critic supplies a fixed tail
value, while the policy is the only module that changes the deployed controller.

The execution objective is defined on a 0.4\,s segment: at 50\,Hz the policy
executes a 20-step block, and the objective is evaluated on its 10-step
differentiable prefix. The remaining 10 steps are not differentiated step by
step; instead, a
fixed, training-only critic estimates their tail return from the imagined state
at step 10. This value enters the loss and remains differentiable with respect to
the imagined state. The critic is held fixed after initialization; it is not
updated during policy training. The
reference is a local segment reference used only during training to define the
drift target. At deployment the VLA receives neither the reference trajectory nor
the critic. The policy therefore learns to reduce drift within each segment,
rather than fitting a complete closed-loop trajectory to a global reference.

Policy starts are drawn only from A$'$ segments. Reference-library frames are not
used since they are idealized reference states rather than states the robot
actually visited, so they have zero execution drift and do not cover the
off-trajectory or recovery states needed for correction. Closed-loop VLA
trajectories are also excluded because they do not provide an aligned local
reference segment for each visited frame, which means the drift target cannot be
formed. In addition, their drift accumulates over a full episode and cannot be
corrected within a 0.4\,s segment. Closed-loop data is therefore used to train
the world model rather than as policy starts.

A frame from A$'$ is eligible as a policy start if it was visited, its root
offset from the segment reference is at most 0.10\,m, at least 10 frames of
history exist, and the 20-step reference window stays inside the clip. We use
only training clips, sample eligible frames with the training coverage weights,
and read only frames and states, not actions. The 0.10\,m bound is a trust region
because a 10-step window can correct only a few centimeters; a larger offset
would act as a nearly constant error. Frames outside the bound still train the
world model, but they are not policy starts.

\paragraph{Forward pass}
An update is a single differentiable program. It samples a token chunk, decodes
it through the frozen BFM, rolls out the world model, and evaluates an execution
cost on the imagined states. Given a conditioning context $c_t$ and a noise
sample $\varepsilon$, the token chunk is produced by four Euler steps,
\begin{equation}
z^{k+1}=z^{k}+\Delta\tau_k\,v_\theta(z^{k},\tau_k,c_t),
\qquad k=0,\ldots,3,
\label{eq:fm_sampler}
\end{equation}
where $z^k$ is the chunk after step $k$, $v_\theta$ is the flow-matching velocity
field, and $\Delta\tau_k$ is the step size. These four steps produce the 40-step
chunk used by the policy.

The body channels are then quantized to the $1/16$ grid used by the BFM,
\begin{equation}
\tilde z=z+\mathrm{sg}\big(Q(z)-z\big),
\label{eq:ste}
\end{equation}
where $Q(z)$ is grid quantization and $\mathrm{sg}$ is the stop-gradient. The
forward path uses the same quantized tokens as deployment, while the backward
path treats the quantizer as identity. This is a straight-through estimator;
the resulting gradient is therefore biased relative to the true quantized
objective.

Starting from the observed state $s_{t|t}$ and history $O_{t|t}$, the BFM and
world model are unrolled over the execution block. Writing
$\hat s_i\equiv\hat s_{t+i|t}$ and $\hat O_i\equiv\hat O_{t+i|t}$,
\begin{equation}
\begin{aligned}
u_i&=\pi_\psi(\tilde z_i,\hat O_i),\\
d_i&=F_\phi(\hat s_i,u_i,\xi_{t+i}),\\
\hat s_{i+1}&=\mathcal{I}(\hat s_i,d_i),\\
\hat O_{i+1}&=\mathcal U(\hat s_{i+1},\hat O_i,u_i),
\end{aligned}
\label{eq:rollout}
\end{equation}
for $i=0,\ldots,H-1$ with $H=10$. The first line decodes the quantized token
together with the proprioceptive history into the normalized body action
$u_i\in\mathbb{R}^{29}$. The second line predicts the state increment with the
world model and applies the integration operator \(\mathcal{I}\) to obtain the
next imagined state $\hat s_{i+1}$ under that action and the simulation context
$\xi_{t+i}$. The fourth line updates the proprioceptive history with the new state
and action, so the next token is decoded under the same history convention as
deployment. The policy update uses \(H=10\) steps. The fixed critic is evaluated
at step \(H\); its value summarizes the remaining \(K-H=10\) steps.

The per-step cost aggregates normalized capped errors,
\begin{equation}
\begin{aligned}
\ell_{k,i}
&=
\frac{\min(e_{k,i},\,\kappa\,\sigma_k^2)}{\sigma_k^2},\\
\ell_{{\rm st},i}
&=
\frac{\min(e_{{\rm st},i},\,\kappa\,\sigma_{\rm st}^2)}
{\sigma_{\rm st}^2},\\
c_i
&=
w(f_{t+i})\,\Delta t
\left(
\sum_{k\in\mathcal A} w_k\,\ell_{k,i}
+
w_{\rm st}\,\ell_{{\rm st},i}
\right),\\
J
&=
-\sum_{i=0}^{H-1}\gamma^{i}c_i,
\end{aligned}
\label{eq:objective}
\end{equation}
where \(c_i\) is a positive cost, \(J\) is the corresponding return,
\(\Delta t=0.02\)\,s is the control period, \(\gamma\) is the discount factor,
\(\kappa\) is the cap multiplier, and \(\mathcal A\) is the active set of tracking
terms. The seven candidate errors \(e_{k,i}\) cover body position, body
orientation, body linear and angular velocity, root position, root orientation,
and root horizontal velocity; terms with zero weight are inactive. The
transition weight is \(w(f_{t+i})=1.5\) at task transitions and \(1\) elsewhere.
For root position, \(\ell_{k,i}\) uses the same capped form but with a
logarithmic tail beyond the cap, so its gradient does not vanish for large drift.

The world model predicts all state channels, including velocities, because they
are needed for integration. Its velocity channels are less reliable than
positions: at \(H=10\) the joint-velocity correlation is about 0.2, and the root
speed is systematically overestimated. We therefore assign zero cost to angular
velocity, root linear velocity, and yaw rate, and keep only a small
body-linear-velocity weight. For the anti-stall term, we use the 3D root
displacement, including its vertical component.

The anti-stall error targets root-motion stalls. Let \(p_i\in\mathbb{R}^{3}\)
denote the root position at step \(i\), and define
\begin{equation}
\begin{aligned}
m_i
&=
\frac{\|p_i-p_{i-1}\|}{\Delta t},\\
m_i^{\rm ref}
&=
\frac{\|p_i^{\rm ref}-p_{i-1}^{\rm ref}\|}{\Delta t},
\end{aligned}
\label{eq:motion}
\end{equation}
where \(m_i\) and \(m_i^{\rm ref}\) are the actual and reference root speeds.
The anti-stall error is
\begin{equation}
\begin{aligned}
e_{{\rm st},i}
={}&
\left(
\min\{[m_i^{\rm ref}-m_i-\delta]_{+},\,\mu_{\rm st}\}
\right)^2\\
&\times
\mathbf{1}\{m_i^{\rm ref}>m_{\rm gate}\}
\mathbf{1}\{m_i<\rho\,m_i^{\rm ref}\},
\end{aligned}
\label{eq:stall}
\end{equation}
where \(\delta=0.05\)\,m/s, \(m_{\rm gate}=0.1\)\,m/s,
\(\mu_{\rm st}=0.6\)\,m/s, and \(\rho=0.10\). These four scalars are selected
once on validation cells and then kept fixed across tasks. The gate opens only
when the reference root motion is nonzero and the actual root motion falls
below a fraction \(\rho\) of the reference. The anti-stall term has a single
generic form and is shared across tasks. It detects root-motion stalls,
including the final squat in SitSofa. Its largest observed effect is on
SitSofa.

The policy update rolls out \(H=10\) steps. The fixed critic is evaluated at
step \(H\); its value summarizes the remaining \(K-H=10\) steps.

The update minimizes the execution cost together with three groups of
differentiable constraints on the joint targets it induces,
\begin{equation}
\begin{aligned}
\mathcal L={}&-\bar J_{\rm ac}
+\lambda_{\rm lim}\mathcal L_{\rm lim}
+\lambda_{\rm trust}\mathcal L_{\rm trust}
+\lambda_{\rm peak}\mathcal L_{\rm peak},\\[2pt]
\bar J_{\rm ac}={}&J+\beta\,\gamma^{H}\,\mathbf{1}_{\rm alive}(H)\,V_H,
\end{aligned}
\label{eq:joint_loss}
\end{equation}
where \(\mathcal L_{\rm lim}\) penalizes joint-limit violations,
\(\mathcal L_{\rm trust}\) is a same-noise trust-region penalty on the deviation
of the joint targets from the pristine FT policy and the projection-only policy,
and \(\mathcal L_{\rm peak}\) caps the per-joint peak deviation. The scalar
\(\beta\) weights the critic value, and \(\gamma\) is the discount factor. Here
\(V_H=V(y_H)\) is the critic estimate of the return over the remaining \(K-H\)
steps, and \(\mathbf{1}_{\rm alive}(H)\) is the survival indicator at step \(H\).
In our setting \(K=20\) and \(H=10\), so \(V_H\) estimates the remaining
10-step tail return; the target is the detached \(K-H\)-step Monte-Carlo return
from \(y_H\). The vector \(y_H\) stacks the imagined state, the reference state,
their difference, the simulation context, the proprioceptive history, and the
planned token segment at step \(H\). The planned-token features in \(y_H\) are
detached, so the critic gradient does not bypass the imagined state. The critic
is held fixed after initialization; its output remains differentiable with
respect to \(y_H\), and the norm of its gradient is capped relative to that of
\(J\). Thus the first 10 steps receive direct execution gradients, while the last
10 steps enter only through the critic value.

\paragraph{Backward pass}
The policy gradient is obtained by one reverse sweep through the unrolled graph.
We use the positive per-step cost \(c_i\) directly, so the execution return is
\[
J=-\sum_{i=0}^{H-1}\gamma^{i}c_i.
\]
The recursion below differentiates the first-\(H\) execution return \(J\). The
critic tail is differentiated separately at \(y_H\) and accumulated into the same
backward sweep; therefore, for \(J\), the boundary adjoints are
\(\delta_H=\mu_H=0\).

Define the state and history adjoints
\[
\delta_i=\frac{\partial J}{\partial\hat s_i},
\qquad
\mu_i=\frac{\partial J}{\partial\hat O_i},
\]
and the local Jacobians
\[
\begin{aligned}
A_i&=\frac{\partial\hat s_{i+1}}{\partial\hat s_i},&
B_i&=\frac{\partial\hat s_{i+1}}{\partial u_i},\\
C_i&=\frac{\partial u_i}{\partial\hat O_i},&
D_i&=\frac{\partial\hat O_{i+1}}{\partial\hat s_i},\\
E_i&=\frac{\partial\hat O_{i+1}}{\partial\hat O_i},&
F_i&=\frac{\partial\hat O_{i+1}}{\partial u_i}.
\end{aligned}
\]
Here \(A_i\) propagates state sensitivity, \(B_i\) maps action changes to
next-state changes, \(C_i\) maps history changes to action changes, \(D_i\) maps
state changes to history changes, \(E_i\) propagates history sensitivity, and
\(F_i\) captures the direct action-to-history path. The last term is needed
because the history update also consumes the action, not only the state and the
previous history.

With \(\delta_H=\mu_H=0\) for \(J\), the adjoints satisfy, for
\(i=H-1,\ldots,0\),
\begin{equation}
\begin{aligned}
\delta_i&=
-\gamma^{i}\frac{\partial c_i}{\partial\hat s_i}
+A_i^{\top}\delta_{i+1}
+D_i^{\top}\mu_{i+1},\\
\mu_i&=
C_i^{\top}\left(
-\gamma^{i}\frac{\partial c_i}{\partial u_i}
+B_i^{\top}\delta_{i+1}
+F_i^{\top}\mu_{i+1}
\right)
+E_i^{\top}\mu_{i+1}.
\end{aligned}
\label{eq:adjoint}
\end{equation}
The first equation combines the discounted direct cost sensitivity to the state
with the future state and history sensitivities. The second does the same for the
history, converts action sensitivity into history sensitivity through \(C_i\),
and includes the direct action-to-history path through \(F_i\). The position-difference
motion estimate makes \(c_i\) depend on \(\hat s_{i-1}\) as well; the implementation
differentiates this dependence through the full graph, while the recursion above
omits it for readability.

The token gradient follows by projecting the action adjoint through the BFM,
\begin{equation}
\frac{\partial J}{\partial\tilde z_i}
=
\left(\frac{\partial u_i}{\partial\tilde z_i}\right)^{\!\top}
\left[
-\gamma^{i}\frac{\partial c_i}{\partial u_i}
+B_i^{\top}\delta_{i+1}
+F_i^{\top}\mu_{i+1}
\right].
\label{eq:token_grad}
\end{equation}
The matrix \(B_i\) is the action Jacobian of the integrated transition at step
\(i\):
\[
B_i=
\frac{\partial \mathcal{I}(\hat s_i,d_i)}{\partial d_i}
\frac{\partial F_\phi(\hat s_i,u_i,\xi_{t+i})}{\partial u_i}.
\]
It therefore includes the integration step, not only the raw output of
\(F_\phi\). For the residual action head in \eqref{eq:bilinear}, it also includes
both the MLP action derivative and \(G_\phi(\hat s_i)\), not \(G_\phi\) alone.

Since the straight-through quantizer has unit derivative, the gradient reaching
the action-head parameters is
\begin{equation}
\nabla_\theta J
=
\sum_{i=0}^{H-1}
\left(\frac{\partial z_i}{\partial\theta}\right)^{\!\top}
\frac{\partial J}{\partial\tilde z_i}.
\label{eq:bptt}
\end{equation}
The sampler Jacobian is expanded as
\begin{equation}
\begin{aligned}
\frac{\partial z_i}{\partial\theta}
&=\sum_{k=0}^{3}P_k\,\Delta\tau_k
\frac{\partial v_k}{\partial\theta},\\
P_k&=\prod_{m=k+1}^{3}\left(I+\Delta\tau_m
\frac{\partial v_m}{\partial z_m}\right),
\end{aligned}
\label{eq:sampler_grad}
\end{equation}
where \(P_k\) is the chunk-level sampler Jacobian and propagates the sensitivity of
the later sampler steps back to step \(k\); \(\partial z_i/\partial\theta\) denotes
its \(i\)-th token block. Only \(\theta\), the action-head parameters (low-rank adapters and the
output projection), receive the policy update; \(\psi\) and \(\phi\) are
untouched. The critic is fixed. Its tail value adds a boundary gradient at \(y_H\), which
is included in the same backward sweep; no critic parameters are updated.

The world model is trained once, on excitation replay together with the
closed-loop episodes collected with the initial policy, and its coefficients are
fixed while the policy is updated. Evaluation always runs in the simulator.
Algorithm~\ref{alg:difftrain} summarises one update, whose cost is four DiT passes
for the sampler, one BFM and one world-model pass per rollout step, one fixed
critic evaluation, and a single backward sweep whose activation footprint grows
with \(H\) rather than with \(K\).

\begin{algorithm}[t]
\caption{One execution-aware differentiable training update.}
\label{alg:difftrain}
\small
\begin{algorithmic}[1]
\Require window \((s_t,O_t,\xi_t)\), context \(c_t\)
\State sample \(\varepsilon\); unroll the sampler (4 steps)
\State \(\tilde z\leftarrow Q(z)\) forward, identity backward
\For{\(i=0\) to \(H-1\)}
  \State \(u_i\leftarrow\pi_\psi(\tilde z_i,\hat O_i)\)
  \State \(d_i\leftarrow F_\phi(\hat s_i,u_i,\xi_{t+i})\)
  \State \(\hat s_{i+1}\leftarrow\mathcal{I}(\hat s_i,d_i)\)
  \State \(\hat O_{i+1}\leftarrow\mathcal U(\hat s_{i+1},\hat O_i,u_i)\)
\EndFor
\State \(V_H\leftarrow V(y_H)\) \Comment{fixed critic; \(y_H\) keeps its graph}
\State \(J\leftarrow-\sum_{i<H}\gamma^{i}c_i\)
\State \(\bar J_{\rm ac}\leftarrow J+\beta\gamma^{H}\mathbf{1}_{\rm alive}(H)V_H\)
\State form \(\mathcal L\)
\State backpropagate; update \(\theta\) \Comment{critic frozen}
\end{algorithmic}
\end{algorithm}
\section{Experiments}
\label{sec:experiments}

\subsection{Experimental Setup}

\textbf{Platforms and tasks.}
We evaluate on HumanoidArena in Isaac Lab simulation with a Unitree G1,
using 29 body degrees of freedom and 14 finger joints at a 50\,Hz body control
rate, 40-step VLA chunks and a 20-step replanning horizon. The simulation task
set contains four HumanoidArena tasks: OpenDoor, SitSofa, Boxing, and VisionNavi.
The main comparison uses these four tasks with a complete baseline grid; all
quantitative closed-loop numbers in this paper come from the simulator.

\textbf{Data and splits.}
Training data are 400 recorded HumanoidArena episodes carrying SONIC motion
tokens (100 per task). Scene instances and instructions are separated before
temporal windowing, so held-out scenes and paraphrases never appear in training.
The VLA is fine-tuned on these episodes. The world model uses the four-source mix
of \Cref{sec:world_model} (clean A$'$, token-perturbed A$'$, closed-loop pairs,
and branch groups), is trained once, and is held fixed during execution-aware
differentiable training;
future states are used only to construct its regression targets and are
unavailable at deployment.

\textbf{Metrics and protocol.}
All success comparisons use the same protocol: four tasks, seeds $0$--$2$, and
repeats $0$--$19$, giving 60 episodes per task and 240 per arm, with matched
episode seeds across arms. Task success is the primary metric. A cell is
successful only if its executed episode succeeded. Failed or timed-out
episodes remain in every denominator. Fall is an episode terminated
by a fall. Paired cells are tested with McNemar's test. Two arms are
compared: the task fine-tuned VLA with a frozen SONIC decoder (\emph{Base}),
and the same model with LoRA adapters trained by execution-aware
differentiable training (\emph{Full}); they share checkpoints, seeds, and
episodes. The Full arm uses the same architecture for all tasks; the base has
no adapter.

\textbf{Training recipe.}
The architecture and training recipe are shared by all tasks (\Cref{tab:recipe}).
The VLA is the GR00T~N1.7 checkpoint fine-tuned on HumanoidArena tokens for
8k steps. LoRA is applied to the attention and feed-forward blocks and the output
projection (rank 16, $\alpha=32$, 16.9M trainable parameters, 1.0\% of the
1.64B-parameter model). The world model is trained once and held fixed. Each
update uses a 40-step chunk, a 20-step rollout, a 10-step differentiable window,
and four Euler sampler steps. Training runs for 300 updates with batch size 32
and a cosine learning rate from $10^{-4}$, on two RTX 4090 GPUs.

\begin{table}[t]
\centering
\footnotesize
\setlength{\tabcolsep}{4pt}
\renewcommand{\arraystretch}{1.05}
\caption{\textbf{Shared training recipe.} The anti-stall term has a single
generic form and is shared across tasks; it is defined on the 3D root
displacement and uses a reference-relative stall gate. All other components are
shared.}
\label{tab:recipe}
\begin{tabular}{@{}ll@{}}
\toprule
VLA & GR00T N1.7, 8k-step task fine-tune \\
Adaptation & LoRA rank 16, $\alpha=32$, 16.9M trainable \\
Data & 400 episodes \\
World model & action-conditioned, trained once, fixed \\
Critic & fixed after initialization \\
Chunk / rollout & 40 / 20 steps \\
Horizon / sampler & $H=10$ / 4 Euler steps \\
Updates / batch / lr & 300 / 32 / $10^{-4}$ cosine \\
$\lambda_{\rm lim}$ / $\beta$ & 0.01855 / 0.975 \\
Hardware & 2$\times$ RTX 4090 \\
\bottomrule
\end{tabular}
\end{table}

\subsection{Closed-Loop Benchmark Comparison}

\begin{figure*}[t]
\centering
\includegraphics[width=\textwidth]{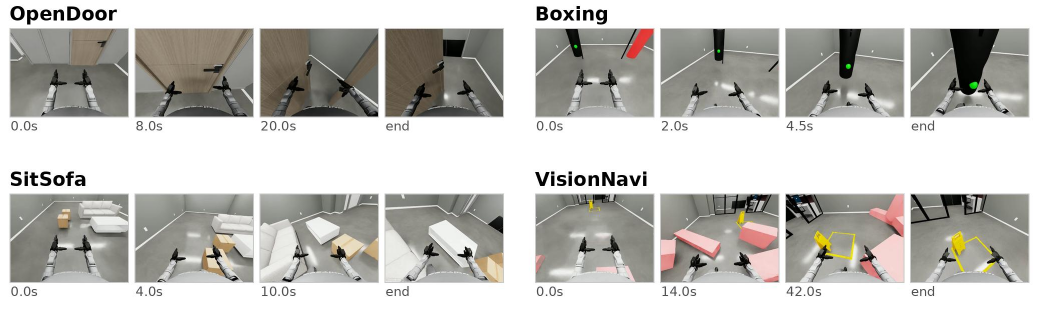}
\caption{\textbf{Simulation evaluation tasks.} Each cell is one task; the frames
show the first frame, two key moments, and the final frame (``end''), labeled by
time or ``end''. OpenDoor: approach, handle contact, and opening. Boxing:
approach and two punch/contact moments. SitSofa: approach, contact, and seated
posture. VisionNavi: crossing the obstacle field to the goal. Frames are
illustrative, not success annotations.}
\label{fig:tasks}
\end{figure*}

We first compare complete closed-loop control. Differentiable training raises average
success by 12.5 points (\Cref{tab:main}), and every task improves; the gain
varies from 10.0 to 16.7 points, with the largest absolute gain on the
contact-rich SitSofa task. We use a mild coverage reweighting so that walking
frames are not under-represented in the start distribution; the ablation below
isolates its effect.

\begin{table}[t]
\centering
\footnotesize
\setlength{\tabcolsep}{2.6pt}
\renewcommand{\arraystretch}{1.05}
\caption{\textbf{Closed-loop results in simulation.} Seeds $0$--$2$ $\times$
repeats $0$--$19$, giving 60 episodes per task and 240 per arm. Fall is the
percentage of episodes terminated by a fall. Entries show Base$\to$Ours; a
single Fall value means both arms are 0.0\%.}
\label{tab:main}
\begin{tabular}{@{}lcccc@{}}
\toprule
Task & Base & Ours & $\Delta$ & Fall \\
     & (\%) & (\%) & (pp) & (\%) \\
\midrule
OpenDoor & 65.0 & 78.3 & $+13.3$ & 0.0 \\
Boxing   & 71.7 & 81.7 & $+10.0$ & 0.0 \\
SitSofa  & 48.3 & 65.0 & $+16.7$ & 0.0 \\
VisionNavi  & 51.7 & 61.7 & $+10.0$ & 15.0$\to$8.3 \\
\midrule
Average  & 59.2 & 71.7 & $+12.5$ & 3.8$\to$2.1 \\
\bottomrule
\end{tabular}
\end{table}

\Cref{tab:main} reports the main comparison. Average success rises from 59.2\% to
71.7\% ($+12.5$ points), and every task improves: on SitSofa success
rises from 48.3\% to 65.0\% ($+16.7$ points), on OpenDoor from 65.0\% to 78.3\%
($+13.3$ points), and on Boxing and VisionNavi from 71.7\% to 81.7\% and from
51.7\% to 61.7\% ($+10.0$ points each). The anti-stall contribution is isolated
in \Cref{tab:leaveoneout}: removing it leaves OpenDoor, Boxing, and VisionNavi
at their Full success rates, while SitSofa drops from 65.0\% to 50.0\%. Falls
are observed only on VisionNavi and are reduced from 15.0\% to 8.3\%
while success there rises by ten points, so the gain cannot be attributed to
riskier behaviour. Failed and timed-out episodes are retained in every
denominator. The aggregate gain is significant on the 240 paired cells (McNemar
$p<0.01$).

\begin{figure}[t]
\centering
\includegraphics[width=\columnwidth]{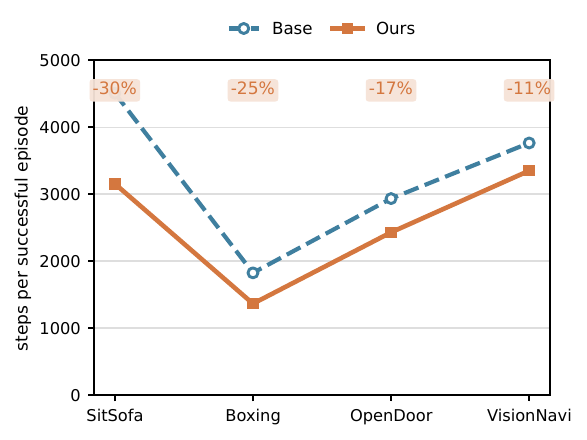}
\caption{\textbf{Cost of a successful episode.} Simulator steps per successful
episode, i.e.\ the total steps of the 60 episodes divided by the number of
successes. Percentages give the change from base to ours; the Full policy is
cheaper on every task.}
\label{fig:cost}
\end{figure}

\Cref{fig:cost} compares the two arms on the number of simulator steps spent per
successful episode, obtained by summing the steps of all 60 episodes and dividing
by the number of successes. The Full policy needs fewer steps on every task: the
cost falls from 2935 to 2431 steps on OpenDoor ($-17\%$), from 3765 to 3351 on
VisionNavi ($-11\%$), from 4508 to 3156 on SitSofa ($-30\%$) and from 1826 to
1366 on Boxing ($-25\%$), with the average decreasing from 3102 to 2490
($-20\%$). The largest savings are obtained where the baseline times out most
often, on SitSofa and Boxing. The extra successes are therefore not bought with
extra interaction time: the same budget yields more successes, and each success
costs less.

\subsection{Surrogate and Gradient Analysis}

Connectivity is necessary. The execution objective alone produces finite,
non-zero gradients on the flow head, all four sampler steps receive gradient, and
the world model's input Jacobian remains active while its parameters are frozen.

We use the surrogate as a direction, not as a calibrated return. Two checks
support this. First, its ranking of counterfactual action sequences over 405
groups is well above chance (70.4\% four-way, $\tau=0.67$ at $H=10$, chance
25\%). Second, in a $\pm\delta$ probe on 87 held-out windows ($H=10$,
$\delta=0.03$), moving along the world-model gradient improves the true return by
$+0.014$ on average, and the predicted and true return changes correlate at
$0.689$. Checkpoints are selected on held-out validation windows using the
differentiable execution return. Among sampling-gradient estimators, the pathwise one keeps token error
at 0.115, the discrete adjoint matches it in return but not in fidelity, and
truncation is biased and raises token error to 0.174.

\begin{table}[t]
\centering
\footnotesize
\setlength{\tabcolsep}{4pt}
\renewcommand{\arraystretch}{1.05}
\caption{\textbf{Gradient-estimator fidelity.} Token error after the same
estimator probe; the return column indicates whether the estimator matches the
pathwise return.}
\label{tab:pathwise}
\begin{tabular}{@{}lcc@{}}
\toprule
Estimator & Token error & Return \\
\midrule
Pathwise & 0.115 & reference \\
Discrete adjoint & 0.121 & matched \\
Truncated & 0.174 & biased \\
\bottomrule
\end{tabular}
\end{table}

\subsection{Choice of the Differentiated Horizon}
\label{sec:horizon}

The horizon $H$ in \eqref{eq:objective} determines how much of the 20-step
execution block receives credit and how far the update trusts the surrogate. A
short window is easier to assign credit to but leaves most of the block
unassigned; a long window covers more of the block but accumulates surrogate
bias. We therefore select $H$ with the surrogate, using two criteria that do not
involve closed-loop success: on counterfactual groups the surrogate must select
the action sequence that the simulator also prefers in at least 70\% of four-way
comparisons (chance 25\%), and its ordering of the four candidates must reach a
Kendall $\tau$ of at least 0.65. Among the windows that satisfy both we take the
\emph{largest}, so that the gradient covers as much of the execution block as the
surrogate can support. This is a ranking criterion, not an absolute-accuracy
criterion: the surrogate is used as a direction signal, not as a calibrated
simulator.

The ranking test uses 405 counterfactual groups collected from body-controller
rollouts driven by perturbed motion tokens. Each group starts from a shared
state and contains four candidate action sequences; the surrogate and the
simulator assign a return to each sequence over the same $H$ steps, and the
resulting orderings are compared. As \Cref{tab:horizon} reports, $H=1$ falls
below both criteria (62.7\%, $\tau=0.56$). Accuracy peaks at $H=5$ (76.8\%,
$\tau=0.73$) and remains above both criteria at $H=10$ (70.4\%, $\tau=0.67$),
but falls below them at $H=20$ (64.2\%, $\tau=0.60$).

\begin{table}[t]
\centering
\footnotesize
\setlength{\tabcolsep}{4pt}
\renewcommand{\arraystretch}{1.05}
\caption{\textbf{Choice of the differentiated horizon.} Four-way selection
accuracy and Kendall $\tau$ between surrogate and simulator returns for each
candidate window. The last column marks the two criteria
($\ge70\%$ and $\tau\ge0.65$).}
\label{tab:horizon}
\begin{tabular}{@{}lccc@{}}
\toprule
$H$ & 4-way acc.\ (\%) & Kendall $\tau$ & Pass \\
\midrule
1  & 62.7 & 0.563 & \\
2  & 72.3 & 0.700 & \checkmark \\
5  & 76.8 & 0.732 & \checkmark \\
10 & 70.4 & 0.669 & \checkmark \\
20 & 64.2 & 0.596 & \\
\bottomrule
\end{tabular}
\end{table}

\begin{figure}[t]
\centering
\includegraphics[width=\columnwidth]{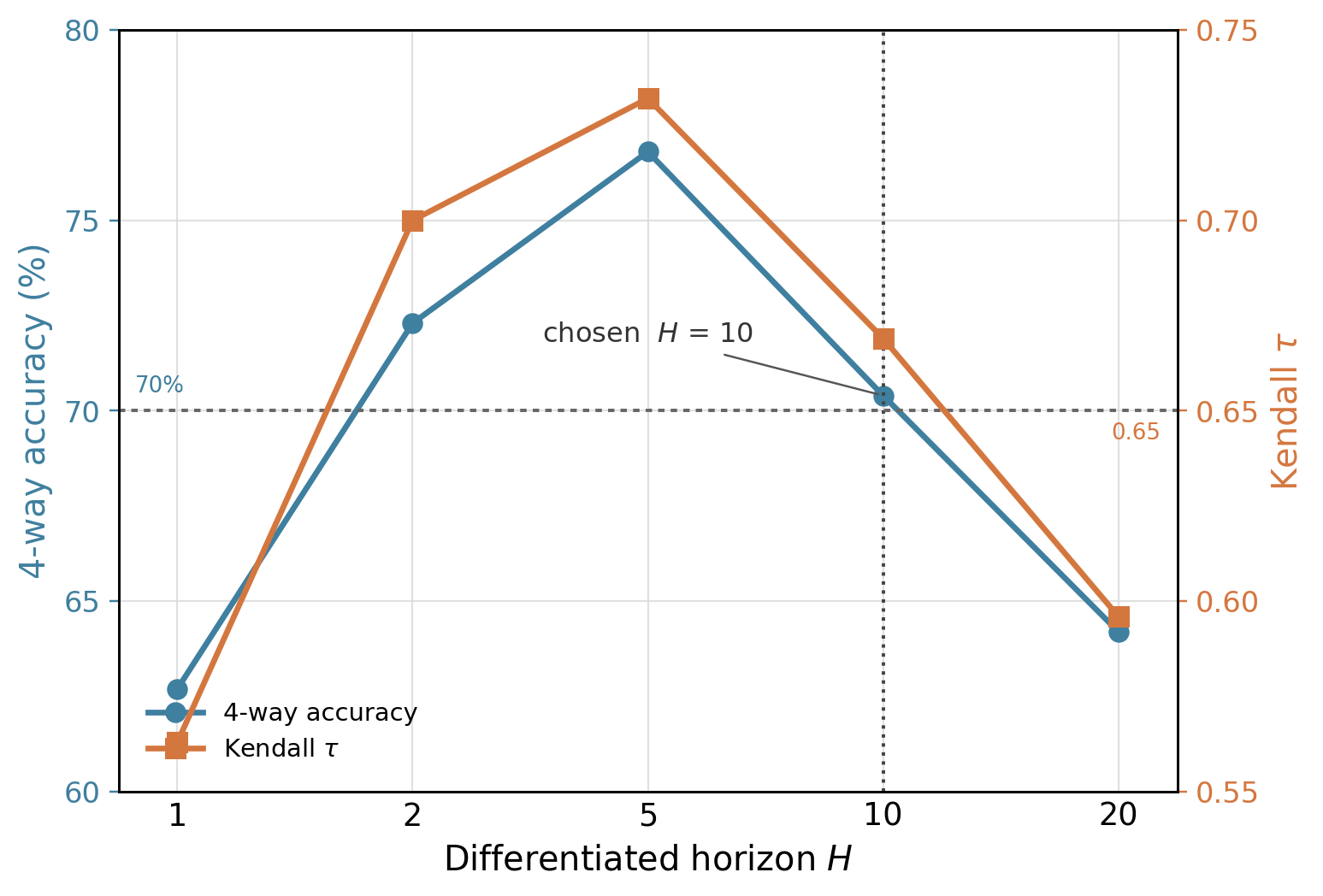}
\caption{\textbf{Surrogate ranking versus horizon.} Four-way selection accuracy
(circles) and Kendall $\tau$ (squares) between surrogate and simulator returns
over 405 counterfactual groups; dashed lines are the thresholds used to choose
the window (70\% and 0.65). The chosen window is $H=10$. The \(H\) values are
shown at equal spacing for readability.}
\label{fig:horizon}
\end{figure}

$H=10$ is the largest window that passes both criteria. $H=5$ ranks best on the
same test but assigns credit to only a quarter of the execution block, whereas
$H=20$ covers the whole block while falling below both criteria. We therefore
fix $H=10$ for every differentiable-training arm; the closed-loop result is a
confirmation at the selected horizon rather than a horizon sweep.

All closed-loop numbers use the same architecture and training pipeline. The
anti-stall term has a single generic form and is active for every task; its
reference-relative gate concentrates the effective gradient on SitSofa. The
architecture, data, checkpoint, horizon, rollout, and optimizer are shared. At $H=10$ the final recipe reaches 78.3\%, 81.7\%, 65.0\%,
and 61.7\% on the four tasks, an average of 71.7\% against 59.2\% for the base
(\Cref{tab:main}).

\subsection{Ablation Studies}

This section reports two complementary sets of ablations. First, we remove one
policy-objective component at a time from the canonical Full recipe and compare
task-level success; Base is a reference. This isolates a gated anti-stall term
from two shared terms, coverage and transition weighting. Second, we vary the
world-model data mix and compare the resulting models on ranking, closed-loop
root error, position action effect, and joint action gain. The two parts answer
different design questions: which objective terms matter at deployment, and
which data sources make the surrogate useful for the gradient.

\subsubsection{Policy-Objective Components}

We remove one component at a time from the canonical Full recipe and compare
task-level success; Base is shown as a reference. The table separates the gated
anti-stall term from the two shared objective terms, so it exposes both task
locality and cross-task effects.

\begin{table}[t]
\centering
\footnotesize
\setlength{\tabcolsep}{3pt}
\renewcommand{\arraystretch}{1.05}
\caption{\textbf{Leave-one-out success on the four tasks.} Full is the canonical
recipe; Base is a reference. Anti-stall is active for every task, but its
effect is concentrated on SitSofa; coverage and transition are shared.
VisNavi abbreviates VisionNavi.}
\label{tab:leaveoneout}
\begin{tabular}{@{}lccccc@{}}
\toprule
Configuration & OpenDoor & Boxing & SitSofa & VisNavi & Avg \\
 & (\%) & (\%) & (\%) & (\%) & (\%) \\
\midrule
Base (reference) & 65.0 & 71.7 & 48.3 & 51.7 & 59.2 \\
Full $-$ coverage & 73.3 & 75.0 & 56.7 & 55.0 & 65.0 \\
Full $-$ transition & 71.7 & 76.7 & 58.3 & 56.7 & 65.8 \\
Full $-$ anti-stall & 78.3 & 81.7 & 50.0 & 61.7 & 67.9 \\
Full & 78.3 & 81.7 & 65.0 & 61.7 & 71.7 \\
\bottomrule
\end{tabular}
\end{table}

\paragraph{Coverage}
Coverage has a broad locomotion signature. Removing it lowers OpenDoor from
78.3\% to 73.3\%, Boxing from 81.7\% to 75.0\%, SitSofa from 65.0\% to 56.7\%,
and VisionNavi from 61.7\% to 55.0\%.

\paragraph{Transition}
Transition weighting is the broadest. Removing it lowers OpenDoor from 78.3\% to
71.7\%, Boxing from 81.7\% to 76.7\%, SitSofa from 65.0\% to 58.3\%, and
VisionNavi from 61.7\% to 56.7\%. The effect is strongest at the handle contact,
the sitting onset, and the approach-to-contact transition, which is why OpenDoor
and SitSofa move by about 6--7 points while Boxing and VisionNavi move by
5 points. The sustained-sitting diagnostic is consistent: with transition
weighting, 11/56 episodes sit continuously, versus 4/56 without (20\% versus
7\%; McNemar $p=0.092$), which we treat as directional rather than significant
evidence.

\paragraph{Anti-stall}
The anti-stall term is active for every task. Its gate uses the 3D root
displacement: it opens when the reference root motion is nonzero and the actual
root motion falls below a fraction \(\rho\) of the reference. Removing the
term leaves OpenDoor, Boxing, and VisionNavi at their Full success rates, while
SitSofa drops from 65.0\% to 50.0\%, and the row remains slightly above Base at
48.3\% because the gradient path and the other objective terms are still present.
The largest observed effect is on SitSofa; per-task gate activation statistics
are not reported.

\paragraph{Overall pattern}
Every intermediate row lies between Base and Full on every task, and no component
explains the entire four-task gap. The rows differ in where they act: coverage is
locomotion-heavy, transition weighting is the most uniform, and anti-stall has
a SitSofa-dominant effect. The gaps are task-specific rather than uniform, consistent with the different
roles of the three terms.

\subsubsection{World-Model Data Sources}

The world-model data mix is a separate ablation axis from the policy-objective
components above. We select it on multiple proxies rather than on ranking alone
(\Cref{tab:wmsources}). Closed-loop pairs are the only source that calibrates the
deployment distribution: H=10 root error falls from 2.1\,cm to 0.45\,cm. Token
perturbation raises the position action-effect $R^2$ from 0.00 to 0.16. Branch
groups improve the joint action sensitivity, raising the joint gain from 0.604 to
0.634, and add a small ranking gain (0.710 to 0.717). The full mix is not the
ranking maximum---A$'$+token ranks 2.8 points higher---but it is Pareto-optimal
on the four proxies and is within 0.02\,cm and 0.02 $R^2$ of the best CL root and
position action $R^2$. We therefore treat it as a balanced design choice rather
than a claim of universal superiority; the production WM adds the SitSofa
windows on top of this mix.

\begin{table}[t]
\centering
\footnotesize
\setlength{\tabcolsep}{3pt}
\renewcommand{\arraystretch}{1.05}
\caption{\textbf{World-model data sources.} All arms share the init, stages, and
420-group ranking. CL root is H=10 on 4,000 held-out closed-loop windows;
position action $R^2$ is H=10 on 400 branch groups; joint gain is the H=1 slope
between predicted and simulator joint action effects. The production WM adds the
SitSofa windows and is not this controlled full arm.}
\label{tab:wmsources}
\begin{tabular}{@{}lcccc@{}}
\toprule
Arm & 4-way & CL root & Pos. $R^2$ & Joint gain \\
 & (\%) & (cm) & & \\
\midrule
Clean A$'$ only & 73.8 & 2.14 & 0.00 & 0.619 \\
A$'$+token & 74.5 & 2.16 & 0.16 & 0.626 \\
A$'$+token+CL & 70.2 & 0.45 & 0.19 & 0.599 \\
no-branch & 71.0 & 0.46 & 0.20 & 0.604 \\
full & 71.7 & 0.47 & 0.18 & 0.634 \\
\bottomrule
\end{tabular}
\end{table}

\subsection{Real-Robot Closed-Loop Inference}
\label{sec:real_robot}

\begin{figure*}[t]
\centering
\includegraphics[width=\textwidth]{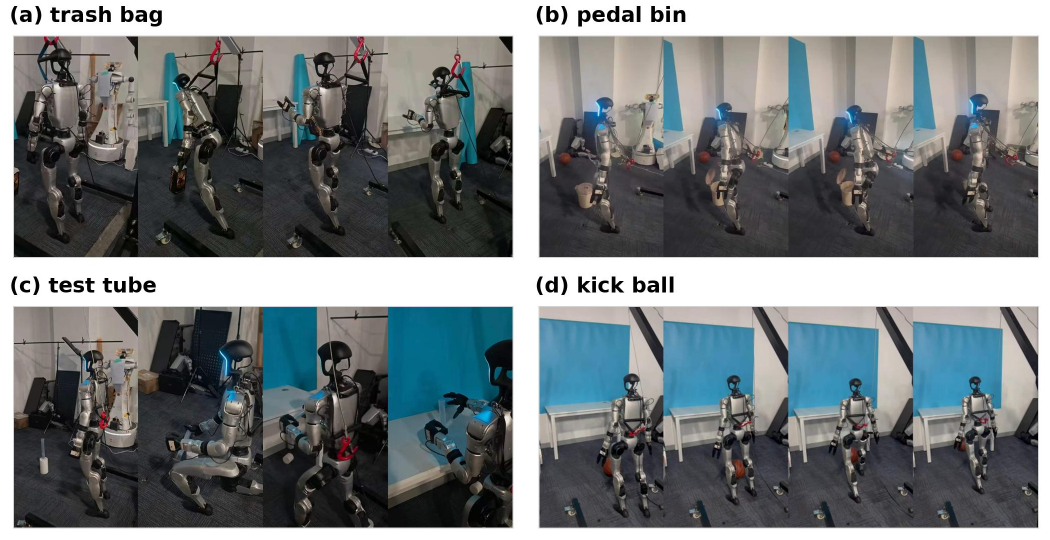}
\caption{\textbf{Real-robot closed-loop inference on a Unitree G1.}
(a) Bend down and pick up a trash bag.
(b) Walk to a pedal-operated bin, press the pedal, and deposit the trash.
(c) Kneel, pick up a test tube, and place it on the table.
(d) Kick a ball to a target position.
Each panel shows the full recorded sequence; frames are illustrative, and no
quantitative success rate is claimed.}
\label{fig:real}
\end{figure*}

We additionally perform real-robot data collection and deployment on a physical
Unitree G1 at 50\,Hz with 20-step replanning. We first collect demonstrations on
the target scenes, re-fine-tune the VLA on these real episodes, and then run the
same execution-aware differentiable training recipe with the real episodes
included in the training mix.
The adapted policy is deployed for closed-loop inference; this is therefore an
adapted real-robot run, not a zero-shot transfer. These runs are a qualitative
validation and are not included in the success-rate table; the quantitative
comparison remains simulation-only. \Cref{fig:real} shows four representative
sequences: (a) pick up a trash bag; (b) press a pedal bin and deposit trash;
(c) place a test tube on a table from a kneeling posture; (d) kick a ball to a
target.

\section{Limitations}
\label{sec:limitations}

Our quantitative evidence is confined to four simulated tasks. The real-robot
runs are qualitative closed-loop inference after real-robot data collection and
adaptation, not a real-robot success-rate benchmark.

The learned world model is a coarse action-conditioned simulator. It rolls out
the execution block and supplies differentiable execution gradients, but it is
not accurate enough to replace the simulator for evaluation or to provide a
calibrated return. On a held-out policy-transition probe, its H=10 error is about
3--4\,cm in root position and 7--8$^\circ$ in rotation on the four-task probe.
This supports coarse credit assignment over the execution window, not accurate
long-horizon prediction. The CL-root column of the data-source table is a
different measurement on held-out closed-loop windows, not the same probe.

The differentiable update is also not a fully on-policy loop: the world model is trained
from a fixed data mix, so the policy cannot track its own
evolving state distribution. This bounds the achievable gain; longer differentiable
training alone is unlikely to close the remaining gap.

The execution objective covers the 29 body joints. The 14 hand channels are
initialized from the 8000-step task fine-tune and are not directly optimized by
the execution objective; we do not evaluate dexterous hand manipulation. Object-centric contact is outside the current state-only world model, which does
not explicitly model object states or contact dynamics.
Generalization to held-out scenes, new object placements, instruction
paraphrases, action delay, observation noise, friction, and external pushes
remains future work.

A natural next step is to improve the world model itself rather than only the
policy optimizer. On the architecture side, velocity-aware prediction heads,
kinematic consistency constraints, and contact-aware state representations could
make the velocity channels accurate enough to enter the execution cost instead
of being downweighted. On the data side, improving the coverage and balance of
the existing transition sources could better represent recovery and contact
states. Better calibration and periodic refresh against the current policy
distribution would further reduce surrogate error. A more accurate model would
make longer differentiable horizons and larger trust regions safer, so the
execution gradient would rely less on surrogate error. Extending the same
treatment to the 14 hand joints is a separate direction.
\section{Conclusion}
\label{sec:conclusion}

We presented an execution-aware differentiable training framework that couples a
flow-matching humanoid VLA with a frozen whole-body controller through a learned
action-conditioned world model. The key idea is to backpropagate execution costs
through the controller and the four-step sampler into the action head, so that
the policy is optimized for the body it actually controls rather than for tokens
alone. On four HumanoidArena tasks, the $H=10$ recipe raises average success from
59.2\% to 71.7\%  and lowers
the cost per successful episode from 3102 to 2490 steps, with the largest gain on
SitSofa; the architecture is shared, and a gated root-motion anti-stall term is enough to
change contact behavior without changing the model. We also close the inference
loop on a physical Unitree G1 after real-robot data adaptation, qualitatively.
The main limits remain the coarse surrogate, the fixed data mix, and the
body-only gradient; an object-aware world model, on-policy refresh, and
held-out or robustness evaluation are left for future work.

\balance
\bibliographystyle{IEEEtran}
\bibliography{references}
\end{document}